%% file: Main.tex
\documentclass[letterpaper, 10 pt, conference]{template/ieeeconf}
\IEEEoverridecommandlockouts                              

\input{Definitions}

\title{\LARGE \bf Convolutional Bayesian Kernel Inference for 3D Semantic Mapping}

\author{Joey Wilson, Yuewei Fu, Arthur Zhang, Jingyu Song \\
Andrew Capodieci, Paramsothy Jayakumar, Kira Barton, and Maani Ghaffari%
\thanks{DISTRIBUTION A. Approved for public release; distribution unlimited. OPSECEC\#6844.}
\thanks{J. Wilson, Y. Fu, A. Zhang, J. Song, K. Barton, and M. Ghaffari are with the University of Michigan, Ann Arbor, MI 48109, USA. {\tt\small{\{wilsoniv,ywfu,arthurzh\}@umich.edu}},
{\tt\small{\{jingyuso,bartonkl,maanigj\}@umich.edu}}}
\thanks{A. Capodieci is with Neya Systems Division, Applied Research Associates, Warrendale, PA 15086, USA. {\tt\small{acapodieci@neyarobotics.com}}}
\thanks{P. Jayakumar is with the US Army DEVCOM Ground Vehicle Systems Center, Warren, MI 48397, USA. {\tt\small{paramsothy.jayakumar.civ@army.mil}}}%
}

\begin{document}

\maketitle
\thispagestyle{empty}
\pagestyle{empty}

\begin{abstract}
Robotic perception is currently at a cross-roads between modern methods, which operate in an efficient latent space, and classical methods, which are mathematically founded and provide interpretable, trustworthy results. In this paper, we introduce a Convolutional Bayesian Kernel Inference (ConvBKI) layer which \textcolor{black}{learns to perform explicit} Bayesian inference within a depthwise separable convolution layer to maximize efficency while maintaining reliability simultaneously. We apply our layer to the task of \textcolor{black}{real-time} 3D semantic mapping, where we learn semantic-geometric probability distributions for LiDAR sensor information \textcolor{black}{and incorporate semantic predictions into a global map}. We evaluate our network against state-of-the-art semantic mapping algorithms on the KITTI data set, demonstrating improved latency with comparable semantic label inference results. 

\end{abstract}



\input{Intro}
\input{Literature}
\input{Method}
\input{Results}
\input{Conclusion}

\clearpage
{\small
\balance
\bibliographystyle{IEEEtran}
\bibliography{bib/strings-abrv,bib/ieee-abrv,bib/refs}
}

\end{document}

%% file: Definitions.tex
\usepackage{graphicx}
\usepackage{caption, subcaption}
\usepackage{amsmath,amssymb,enumerate}

\usepackage{amsthm}
\usepackage{mathtools}
\usepackage{breqn}
\usepackage[usenames,dvipsnames,svgnames,table, xcdraw]{xcolor}
\usepackage[colorlinks=true,pdfpagemode=UseNone,citecolor=black,linkcolor=black,urlcolor=BrickRed]{hyperref}
\usepackage{algorithm, algpseudocode}

\usepackage{blindtext}
\usepackage{gensymb}
\usepackage{xparse}
\usepackage{lipsum}
\usepackage{mathrsfs}
\usepackage[mathscr]{euscript}
\usepackage{times}
\usepackage[noadjust]{cite} 
\usepackage{multicol}
\usepackage{amsfonts}
\usepackage[utf8]{inputenc}
\usepackage[T1]{fontenc}
\usepackage{textcomp}
\usepackage{balance}
\usepackage{soul}
\usepackage{multirow}
\usepackage{booktabs}
\usepackage{sidecap}
\usepackage{makecell}
\usepackage{array,float}

\renewcommand{\thefigure}{\arabic{figure}}

\definecolor{free}{rgb}{0.9608, 0.5882, 0.3922}
\definecolor{building}{rgb}{0.9608, 0.9020, 0.3922}
\definecolor{barrier}{rgb}{0.5882, 0.2353, 0.1176}
\definecolor{other}{rgb}{0.7059, 0.1176, 0.3137}
\definecolor{pedestrian}{rgb}{1, 0.3137, 0.3922}
\definecolor{pole}{rgb}{0.1176, 0.1176, 1}
\definecolor{road}{rgb}{0.7843, 0.1569, 1}
\definecolor{ground}{rgb}{0.3529, 0.1176, 0.5882}
\definecolor{sidewalk}{rgb}{1, 0, 1}
\definecolor{vegetation}{rgb}{1, 0.5882, 1}
\definecolor{vehicles}{rgb}{0.2941, 0, 0.2941}
\definecolor{fence}{rgb}{0.1961, 0.4706, 1}
\definecolor{sign}{rgb}{0, 0, 1}

\definecolor{scarColor}{rgb}{0.9608, 0.5882, 0.3922}
\definecolor{sbicycleColor}{rgb}{0.2608, 0.9020, 0.3922}
\definecolor{smotorcycleColor}{rgb}{0.5882, 0.2353, 0.1176}
\definecolor{struckColor}{rgb}{0.7059, 0.1176, 0.3137}
\definecolor{sothervehicleColor}{rgb}{1, 0.3137, 0.3922}
\definecolor{spersonColor}{rgb}{0.1176, 0.1176, 1}
\definecolor{sbicyclistColor}{rgb}{0.7843, 0.1569, 1}
\definecolor{smotorcyclistColor}{rgb}{0.3529, 0.1176, 0.5882}
\definecolor{sroadColor}{rgb}{0.5, 0, 1}
\definecolor{sparkingColor}{rgb}{1, 0.5882, 1}
\definecolor{ssidewalkColor}{rgb}{0.2941, 0, 0.2941}
\definecolor{sothergroundColor}{rgb}{0.2941, 0, 0.6863}
\definecolor{sbuildingColor}{rgb}{0, 0.7843, 1}
\definecolor{sfenceColor}{rgb}{0.1961, 0.4706, 1}
\definecolor{svegetationColor}{rgb}{0, 0.6863 ,0}
\definecolor{strunkColor}{rgb}{0, 0.2353, 0.5294}
\definecolor{sterrainColor}{rgb}{0.3137, 0.9412, 0.5882}
\definecolor{spoleColor}{rgb}{0.5882, 0.7412, 1}
\definecolor{strafficsignColor}{rgb}{0, 0, 1}

\newcommand{\m}{\mathop{\mathrm{m}}}

\newcommand{\Hz}{\mathop{\mathrm{Hz}}}



%% file: Intro.tex
\section{Introduction}

Robust world models are essential for safe and reliable autonomous robots. Within a world model, an autonomous robot can embed a high level of scene understanding through multiple modalities of information, such as semantic or motion labels. One common world model is a map, where a geometric framework models the world in a manner interpretable to both robots and humans, encouraging reliability and trust. 

Although some works have proposed to discard maps in lieu of an end-to-end deep learning autonomous robot framework, a world model is still critical for safety and trustworthiness. Through a world model, robot failures can be safely diagnosed post-mortem and understood by humans due to the shared human-robot understanding. 

Semantic mapping is a framework for robotic mapping which extends the geometric map to include scene ontology. Semantic labels incorporate a higher level of scene understanding by labeling the world with semantics, such as people and chairs. This information can be beneficial for robotic behavior planning. 

Recently, works within mapping have explored learning-based neural implicit representations, which move beyond the structured geometric representations of earlier, probabilistic hand-crafted algorithms. Despite efficient latent operations, there is still a clear trade-off. While maps encoded in a latent space are argued to be more efficient, trainable, and faster, they lose the reliability and trustworthy behavior of hand-crafted mapping methods. In contrast, hand-crafted mapping methods are mathematically derived and can be understood with quantifiable uncertainty, which is necessary for predicting robot failures.

In this paper, we attempt to combine the advantages of deep learning-based approaches with the safe nature and predictability of hand-crafted approaches for 3D semantic mapping. Concretely, we demonstrate that a probabilistic Bayesian inference semantic mapping approach \cite{MappingSBKI} \textcolor{black}{can} be written as a differentiable depthwise convolution \cite{MobileNet} layer, thus enabling an end-to-end mapping framework with the efficiency, speed, and trainable nature of deep learning frameworks, while maintaining quantifiable uncertainty and reliability of a hand-crafted approach. Our main contributions are as follows.

\begin{enumerate}[i.]
    \item Create a real-time 3D semantic mapping neural network layer, which finds middle-ground between classical robotic mapping and modern deep learning.
    \item Propose novel differentiable kernels for Bayesian semantic mapping, and demonstrate improved performance through optimization.
    \item Open source all software for future development at \href{https://github.com/UMich-CURLY/NeuralBKI}{https://github.com/UMich-CURLY/NeuralBKI}.
\end{enumerate}

\begin{figure}[t]
    \includegraphics[width=\linewidth]{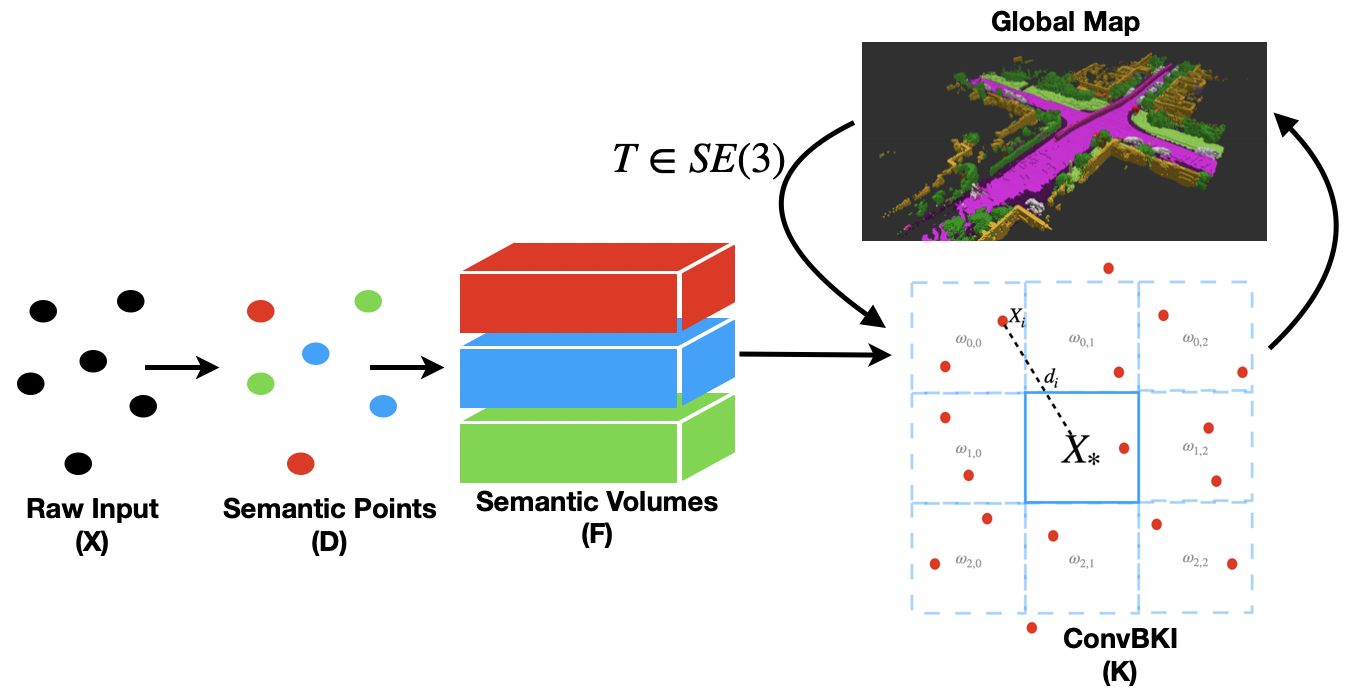}
    \centering
    \caption{Structural diagram of ConvBKI. 3D points are assigned semantic labels from off-the shelf semantic segmentation networks, and grouped into voxels by summing coinciding points. The constructed semantic volumes are convolved with a depthwise filter to perform a real-time Bayesian update on a semantic 3D map. 
    }
    \label{fig:BKINet}
    \vspace{-4mm}
\end{figure}

%% file: Literature.tex
\section{Literature Review}
In this section, we review 3D semantic mapping and the trade-offs between learned and hand-crafted approaches.

\subsection{Learned vs. Hand-Crafted Mapping}
Historically, most mapping methods were hand-crafted and mathematically derived. Early semantic mapping algorithms semantically labeled images, then projected to 3D and directly updated matching voxels through a voting scheme or Bayesian update \cite{SemanticBayes3D, SemanticFrequent3D, SemanticFusion, SemanticVoting3D, SemSegMap}. Later semantic mapping algorithms applied further optimization through Conditional Random Fields (CRF), which encourages consistency between adjacent voxels \cite{SemanticCRF, SemanticCRF2, SemanticCRF3}. Separately, continuous mapping algorithms estimate occupancy through a continuous non-parametric function such as Gaussian processes (GPs) \cite{OccupancyGP, GPOM}. However, these methods suffer from a high computation load, rendering them impractical for on-board robotics. For example, GPs have a cubic computational cost with respect to the number of data points and semantic classes \cite{SemanticGP}. \textcolor{black}{Other works have also explored semantic mapping with alternative data-efficient representations such as surfels \cite{SumaPP}, truncated signed distance functions \cite{PanopticTSDF}, and meshes \cite{SemanticMesh, Kimera}.}

Many modern approaches to mapping take advantage of neural networks to learn an efficient, implicit approximation of the world in a lower dimensional latent space \cite{SemanticMapNet}. Some approaches include applying recurrent neural networks \cite{RecurrentOctoMap, DA-RNN} or spatio-temporal convolution networks \cite{MotionNet, MotionSC} to model spatio-temporal dynamics. Other recent works have explored approximating continuous geometry implicitly with Neural Radiance Fields (NeRF) \cite{BlockNerf, NeRF} or occupancy networks \cite{OccupancyNetworks, NeuralBlox, ConvOccupancy}, in order to negate the expensive memory of voxels.

While learning-based approaches have succeeded in minimizing memory or accelerating inference, they still encounter significant challenges. By implicitly approximating functions, there is no notion of when a network will fail, as provided by variance or the ability to diagnose an error. On the other side of the spectrum, mathematical hand-crafted approaches provide reliability and trustworthiness at the cost of efficiency. 

\subsection{Bayesian Kernel Inference}


Semantic Bayesian Kernel Inference (S-BKI) \cite{MappingSBKI} is a 3D continuous semantic mapping framework which builds on the work of \cite{BKIProof} and \cite{BKIOccupancy}. BKI is an efficient approximation of GPs, requiring $\mathcal{O}(\text{log} N)$ operations and $\mathcal{O}(N)$ memory instead of $\mathcal{O}(N^3)$, where $N$ is the number points. In contexts such as mapping, there may be hundreds of thousands of points, rendering GPs impractical.

For supervised learning problems, our goal is to identify the relationship $p(y | x_*, \mathcal{D})$ given a sequence of $N$ independent observations $\mathcal{D} = \{(x_1, y_1), ..., (x_N, y_N)\}$, where $x_*$ is a query point. $y$ represents observation values drawn from set $Y$ corresponding to input values $x$ drawn from set $X$. In 3D semantic mapping, the likelihood represents a distribution of semantic labels $Y$ over geometric positions $X$. 

Vega-Brown et al.~\cite{BKIProof} introduce a model and constraints which generalize local kernel estimation to Bayesian inference for supervised learning. They show that the maximum entropy distribution $g$ satisfying $D_{KL}(g||f) \geq \rho(x_*, x)$ has the form $g(y) \propto f(y) ^ {k(x_*, x)}$. In this case, $\rho: X \times X \to \mathbb{R}^+$ is some function which bounds information divergence between the likelihood distribution $f(y_i) = p(y_i | \theta_i)$ and the extended likelihood distribution $g(y_i) = p(y_i | \theta_*, x_i, x_*)$. Functions $k$ and $\rho$ have an equivalence relationship, where each is uniquely determined by the other. The only requirements are that: 
\begin{equation}\label{eq:req1}
    k(x, x) = 1 \forall x \quad \text{and} \quad k(x, x') \in [0, 1] \forall x, x',
\end{equation}
\noindent where $k$ is the kernel function. This formulation is especially useful for likelihoods $p(y|\theta)$ chosen from the exponential family, as the likelihood raised to the power of $k(x_*, x)$ is still within the exponential family. 

Doherty et al.~\cite{BKIOccupancy} then apply the BKI kernel model to the task of occupancy mapping. In occupancy mapping, occupied points are measured by a 3D sensor such as LiDAR, and free space samples \textcolor{black}{can} be approximated through ray tracing. Measurement $x_i \in \mathbb{R}^3$ then represents a 3D position with corresponding observation $y_i^c \in \{0, 1\}$, either indicating free space ($y_i^0 = 1$) or occupied space ($y_i^1 = 1$). In this case, $c \in \mathcal{C}$ is a binary variable indicating whether the point is occupied ($c=1$) or free ($c=0$). Adopting a prior distribution Beta($\alpha_0^0, \alpha_0^1$) over $\theta_0$ yields a closed-form update equation at each time step $t$, such that:
\begin{equation}\label{eq:Occupancy Update}
    \alpha^c_{*,t} = \alpha^c_{*, t-1} + \sum_{i=1}^{N_t}k(x_*, x_i) y_i^c ,
\end{equation}
where * is the query voxel with centroid $x_*$ and parameters $\theta_*$. The equation provides a closed-form method for updating the belief that voxel * is occupied or free, given observed measurements and samples of free space. The kernel depends on distance of observed points to the centroid of each voxel, providing more weight to close points. 
Gan et al.~\cite{MappingSBKI} show that the same approach \textcolor{black}{can} be applied to semantic labels by adopting a Categorical likelihood and placing prior distribution Dir($C, \alpha_0$) over $\theta_*$. Semantic labels $y_i$ \textcolor{black}{are} obtained as estimations from \textcolor{black}{state-of-the-art} neural networks. This model is also calculated using \textcolor{black}{Eq. }\eqref{eq:Occupancy Update}, where the variable $c$ is no longer binary, but represents one of $C$ labels. $y_i$ is again a Categorical distribution, representing the probability of each semantic category. From the Dirichlet distribution concentration parameters $\alpha_*$, the expectation and variance of voxel $*$ \textcolor{black}{is calculated as}:
\begin{equation}\label{eq:Variance}
    \eta_*^c = \sum_{j=1}^C \alpha^j_*, \quad \mathbb{E}[\alpha^c_*] = \frac{\alpha^c_*}{\eta_*^c}, \quad \mathbb{V}[\alpha^c_*] = \frac{
    \frac{\alpha^c_*}{\eta_*^c} (1 - \frac{\alpha^c_*}{\eta_*^c}) 
    }{ 1 + \eta_*^c
    }.
\end{equation}



Although Semantic BKI has succeeded in 3D mapping, it is still limited in a few key ways. Firstly, the kernel is hand-crafted, and kernel parameters must be manually tuned. As a result, a single spherical kernel is shared between all semantic classes. Second, the update operation has a slow inference rate, as the kernel evaluation requires a nearest neighbor operation.


%% file: Method.tex
\section{Method}
We propose a novel neural network layer, Convolutional Bayesian Kernel Inference (ConvBKI), which is intended to accelerate and optimize S-BKI. Compared to S-BKI, ConvBKI \emph{learns} a unique kernel for each semantic class, and generalizes to 3D ellipsoids instead of restricting distributions to spheres. We demonstrate how to train the layer and incorporate it into an end-to-end deep neural network for updating semantic maps \textcolor{black}{in static environments}.

\subsection{Convolutional BKI} 
We build a faster, trainable version of Semantic BKI based on the key observation that \textcolor{black}{Eq. }\eqref{eq:Occupancy Update} \textcolor{black}{can} be rewritten as a depthwise convolution \cite{MobileNet}. We find that the kernel parameters are differentiable with respect to a map loss function and \textcolor{black}{are therefore learnable}. Learning the kernel parameters enables more expressive geometric-semantic distributions and improved semantic mapping performance. 



The update operation in \textcolor{black}{Eq. }\eqref{eq:Occupancy Update} performs a weighted sum of semantic probabilities over the local neighborhood of voxel centroid $x_*$. This operation can be directly interpreted in continuous space with radius neighborhood operations such as in PointNet++ \cite{PointNet++}, DGCNN \cite{EdgeConv}, or KPConv \cite{KPConv}. However, we found that in practice, these operations are much too slow to compute for hundreds of thousands of camera or LiDAR points due to an expensive k-Nearest Neighbor operation. Instead, we perform a discretized update, where the geometric position of each local point is rounded to the position of the map voxel it falls in. Approximation through downsampling is already performed in Semantic BKI \cite{MappingSBKI}, and is a common step in real-time mapping literature \cite{VoxBlox, Kimera}. 

Given the prior local map of dimension $\mathbb{R}^{D_C \times D_X \times D_Y \times D_Z}$ and a labeled input point cloud, we first group points within corresponding voxels. $D$ represents the dimension of the \textcolor{black}{semantic} channel ($C$) and Euclidean ($X, Y, Z$) axes. Let $I(*, i)$ be an indicator function representing whether point $x_i$ lies within voxel $*$. From the semantic predictions over each point cloud, we compute input semantic volume $\textbf{F} \in \mathbb{R}^{D_C \times D_X \times D_Y \times D_Z}$ as follows, where input $\textbf{F}_*$ is the sum of all point-wise semantic predictions contained in voxel $*$.

\begin{equation}\label{eq:Input Definition}
    \textbf{F}^c_{*} =  \sum_{i=1}^{N} I(*, i) y_i^c .
\end{equation}

For each voxel, the Bayesian update can be calculated as the sum of the prior semantic map and a depthwise convolution over input $\textbf{F}$. Let $h, i, j$ be the discretized coordinates of voxel $*$ within $\textbf{F}$, and $k, l, m$ be indices within discretized kernel $\mathbf{K} \in \mathbb{R}^{D_C \times f \times f \times f}$ where $f$ is the filter size. Then, we can write the update for a single \textcolor{black}{semantic} channel of voxel $*$ as 
\begin{equation}\label{eq:Depthwise Update}
    \alpha^c_{*, t} = \alpha^c_{*, t-1} + \sum_{k, l, m}  \textbf{K}^c_{k, l, m} \textbf{F}^c_{h+k, i+l, j+m},
\end{equation}

where indices $k, l, m \in [-\frac{f-1}{2}, \frac{f-1}{2}]$. Note that this is the equation for a zero-padded depthwise convolution, where dense 3D convolution is performed at each voxel in the feature map, with a unique kernel $\textbf{K}^c$ for semantic category $c$. As a result, this operation \textcolor{black}{can} be accelerated by GPUs and optimized through gradient descent. 

Following \cite{MappingSBKI} and \cite{BKIOccupancy}, we use a sparse kernel \cite{SparseKernel} as our kernel function since the sparse kernel fulfills the requirements listed in \textcolor{black}{Eq. }\eqref{eq:req1}. Additionally, the sparse kernel is differentiable so that a partial derivative of the loss function with respect to the kernel parameters \textcolor{black}{can} be calculated. The sparse kernel is shown in \textcolor{black}{Eq. }\eqref{eq:SparseKernel}, where the parameters are kernel length $l$, and signal variance $\sigma_0$. Note that for \textcolor{black}{Eq. }\eqref{eq:req1} to remain valid, $\sigma_0$ must be 1, leaving only one tune-able parameter for the kernel function. For two points $x$ and $x'$, let $d := \lVert x - x' \rVert$. The sparse kernel is calculated as 
\begin{align}
\label{eq:SparseKernel}
        \nonumber &k(d) = \\ 
        &\begin{cases}
          \sigma_0 [\frac{1}{3} (2 + \cos (2 \pi \frac{d}{l})(1 - \frac{d}{l}) +
          \frac{1}{2 \pi} \sin (2 \pi \frac{d}{l})) ], 
          & \text{if } d < l \\
          0, & \textcolor{black}{\text{otherwise}}
        \end{cases}.
\end{align}

Effectively, kernel $\mathbf{K}$ is a weight matrix where each weight represents a semantic and spatial likelihood of correlated points. For example, if a point \textcolor{black}{has semantic class road}, then points nearby along the $X$ or $Y$ axes are \textcolor{black}{also likely to have semantic class} road, and would have a high weight. In contrast, a point labeled as pole would have more influence over points nearby vertically rather than horizontally.

While it is possible to learn an individual weight for each position and semantic category in filter $\textbf{K}$, we found that restricting the number of parameters through a kernel function increases the ability of the network to learn generalizable semantic-geometric distributions quickly. Therefore, we learn a sparse kernel $k^c(\cdot)$ for each semantic category, and assign kernel values to $\mathbf{K}$ at each filter index, where distance depends on the resolution $\Delta r$ of the voxel map. For a filter of dimension $f$ and resolution $\Delta r$, the kernel weights $\textbf{K}^c_{k,l,m}$ at filter indices $k, l, m$ \textcolor{black}{are} calculated by evaluating kernel function $k^c$ at the offset of position $k, l, m$ from the centroid as follows. 
\begin{equation}\label{eq:KernelWeight}
    \textbf{K}^c_{k, l, m} = k^c (\lVert \Delta r \cdot (\frac{f-1}{2} - 
    \begin{bmatrix}
        k \\
        l \\
        m
    \end{bmatrix}) \rVert_2)
\end{equation}

\begin{figure}[t]
    \includegraphics[width=0.7\linewidth]{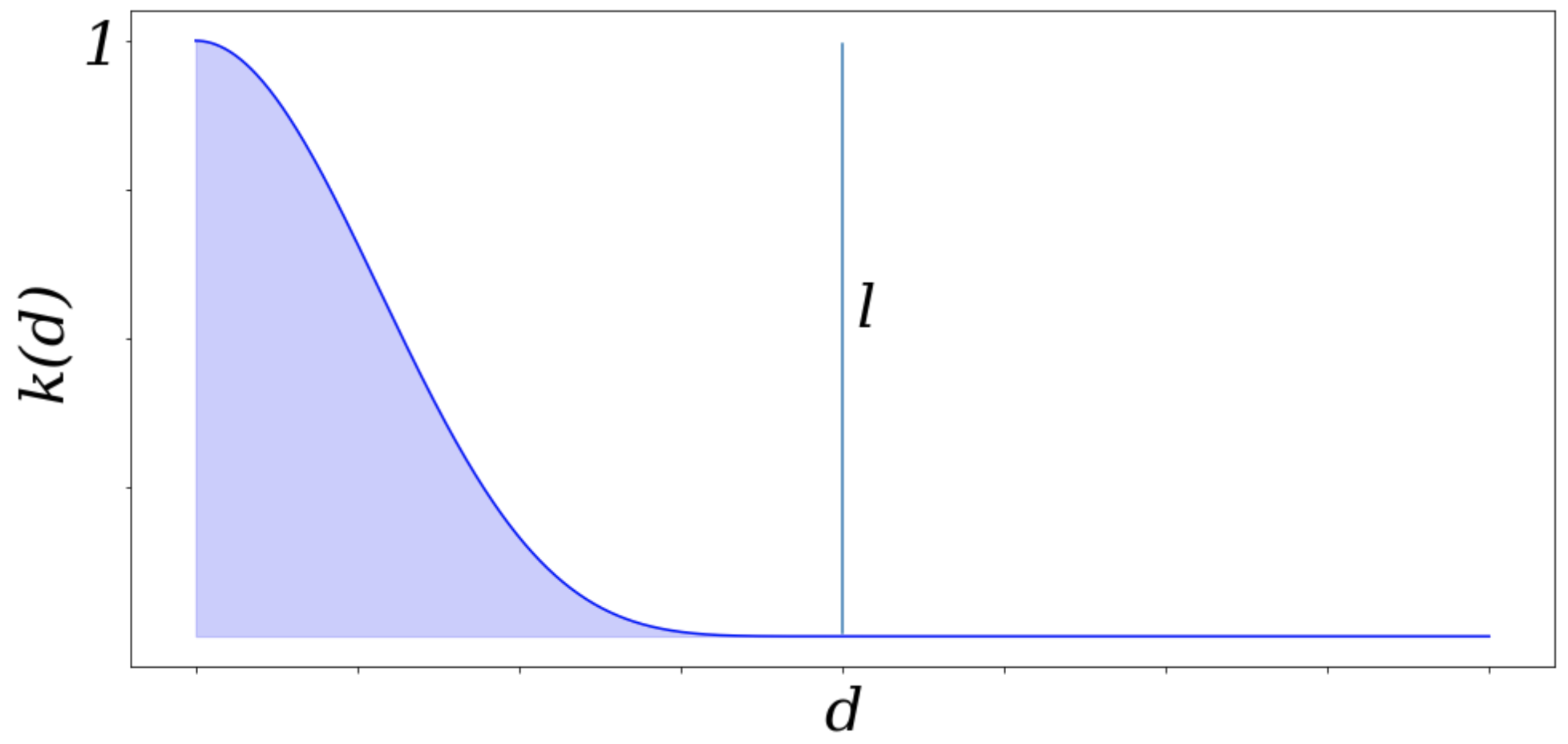}
    \centering
    \caption{Sparse Kernel Function. $k(d)$ has a maximum value of 1 at $d=0$, and decays to 0 by $d=l$. Applied to semantic mapping, points proximal to the voxel centroid have more influence over the semantic label of the voxel.
    }
    \label{fig:SparseKernel}
    \vspace{-4mm}
\end{figure}

A plot of the sparse kernel function is included in Fig. \ref{fig:SparseKernel} for reference. To accommodate complex geometric structures of real objects, we also propose a compound kernel~\cite[Ch. 4]{rasmussen2006gaussian} computed as the product of a kernel over the horizontal plane ($k_h$) and vertical axis ($k_v$) as
\begin{equation}\label{eq:CompoundKernel}
    f(\begin{bmatrix}
        x\\
        y\\
        z\\
    \end{bmatrix} ,
    \begin{bmatrix}
        x'\\
        y'\\
        z'\\
    \end{bmatrix})
    = k_h(\lVert \begin{bmatrix}
        x-x' \\
        y-y' \\
    \end{bmatrix} \rVert) 
    k_v(\lVert \begin{bmatrix}
        z-z' \\
    \end{bmatrix} \rVert).
\end{equation}

\begin{figure}[t]
    \includegraphics[width=0.85\linewidth]{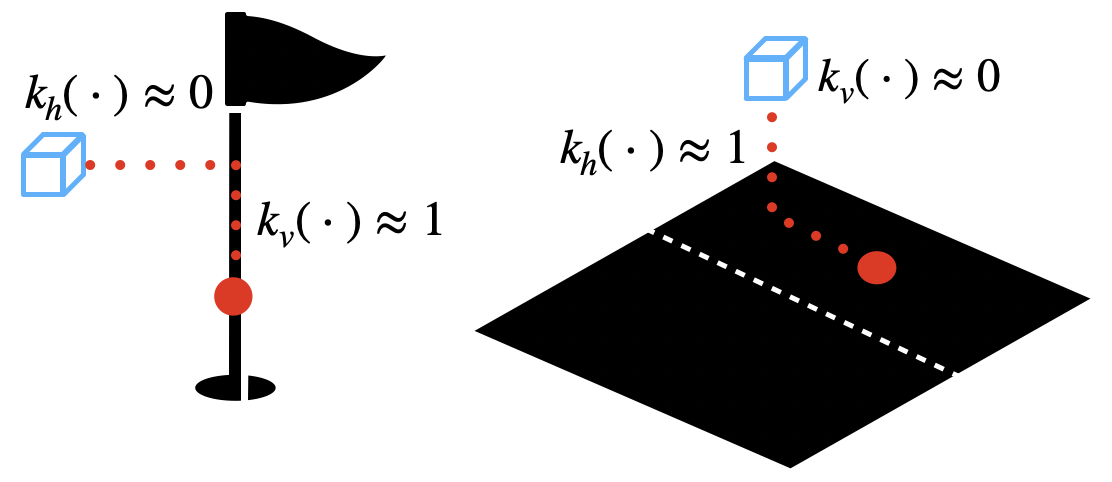}
    \centering
    \caption{Illustration of compound kernel motivation. ConvBKI learns a distribution to geometrically associate points with voxels. Whereas a point (red) labeled pole suggests a vertically adjacent voxel (blue) may also be a pole, it does not imply the same for a horizontally adjacent voxel. Likewise, a point labeled as road suggests horizontally adjacent voxels are also road but not vertically located voxels. Hence, a compound kernel enables ConvBKI to learn more expressive semantic-geometric distributions. 
    }
    \label{fig:ComKernel}
    \vspace{-4mm}
\end{figure}

Intuitively, ConvBKI treats the output of a semantic segmentation neural network as sensor input, and learns a geometric probability distribution over each semantic class. Semantic classes have different shapes, where classes such as poles are more vertical and classes such as road have influence horizontally. The motivation behind a compound kernel for each semantic class is visualized in Fig.~\ref{fig:ComKernel}.

\subsection{Global Mapping}

Next, we apply our ConvBKI layer to the task of global mapping. In global mapping, sensor input is used to construct a full map of the environment, maintaining all past information. 

\textcolor{black}{At initialization, the map consists of an empty set of voxels. Concentration parameters of new voxels in the local region of the ego vehicle are assigned to prior, which is a small non-zero value for each semantic channel.} At each time step, the input to our network is a global pose $T_t \in \text{SE(3)}$, and 3D data $\mathcal{X}_t$ in the form of a point cloud, stereo image, or both. From our prior global map $G_{t-1}$, we query position of $T_t$ to identify the nearest voxel $*$ to the ego position and the local set of voxels with the same shape as $\textbf{F}$. The local voxels serve as a prior local map $L_{t-1}$ for the Bayesian update.

Once we have obtained prior local map $L_{t-1}$, a semantic segmentation network predicts labels $\mathcal{Y}_t$ for 3D points $\mathcal{X}_t$. Next, we apply \textcolor{black}{Eq. }\eqref{eq:Input Definition} to calculate \textcolor{black}{the} input to our ConvBKI layer, aligning points $\mathcal{D}_t$ with local map $L_{t-1}$. Finally, we apply the 3D depthwise convolution from \textcolor{black}{Eq. }\eqref{eq:Depthwise Update} to update local map $L_{t-1}$ and obtain updated states $L_{t}$. The updated voxels from $L_t$ are transferred back to CPU and replace their prior states in $G_{t-1}$ to form $G_t$. 

To accelerate computation and maximize efficiency for the global mapping operation, we make a couple of design choices. First, the global map $G$ is stored on CPU memory \textcolor{black}{and the local update is performed on GPU due to restricted GPU memory.} For efficient retrieval of the local map, the global map is stored in a matrix where each row contains a key and value. The key is the voxel discretized indices, and the value is the semantic concentration parameters. The local map \textcolor{black}{can} be obtained in real-time by batch querying all voxels within local boundaries. We also accelerate runtime by applying garbage collection, where voxels which have not been updated recently (10 frames) are removed from memory to reduce the search space.

\subsection{Training}
We train the kernel functions separately from the respective mapping algorithms for memory efficiency and speed. For static data, applying ConvBKI over each time step individually is equivalent to applying ConvBKI once over all points since the operation is merely a weighted sum. Therefore, when training, we load the past $\mathcal{T}$ point clouds with predicted semantic labels $\mathcal{D}_{t-\mathcal{T}:t}$ and transform all points to the current frame $T_t$. All semantically labeled 3D points are then used to create input encoding $\textbf{F}_t$ through \textcolor{black}{Eq. }\eqref{eq:Input Definition}, so that only one convolution is performed instead of a convolution at each time step. 

\textcolor{black}{ConvBKI learns a probabilistic distribution of semantic segmentation labels over geometrically neighboring predictions. Therefore, ConvBKI must be trained on noisy semantic segmentation predictions similar to the test test. Since semantic segmentation networks achieve higher performance on data they have been trained on, ConvBKI must be trained on a held out set, such as a validation set.} Empirically, training on the validation set instead of training set results in a nearly 4\% improvement in mean Intersection over Union (mIoU) on the test set of Semantic KITTI \cite{SemanticKITTI}. 

%% file: Results.tex
\section{Results}

We perform ablation studies on hyper-parameters of ConvBKI, then compare performance with previous 3D semantic mapping baselines. Lastly, we visualize the semantic-geometric distributions learned by the ConvBKI layer. For each set of results, we compare ConvBKI with a single kernel shared between all semantic classes (ConvBKI Single), ConvBKI with one kernel for each semantic category (ConvBKI Per Class), and ConvBKI with a compound kernel for each semantic category (ConvBKI Compound). 

\textcolor{black}{We compare against two versions of S-BKI to enable direct comparison. S-BKI with 0.2 $\m$ resolution and discretization is a direct comparison to our work, equivalent to ConvBKI Single without optimization at a kernel length of 0.3 $\m$. We also compare against the reported results of S-BKI from \cite{MappingSBKI}, which runs without discretization at a voxel resolution of 0.1 meters and with tuned thresholding. We refer to the S-BKI baselines as S-BKI (0.2m) and S-BKI (fine) where fine indicates a 0.1 $\m$ resolution compared to our 0.2 $\m$ resolution without discretization. S-BKI reports a latency of 2 $\Hz$ with downsampling and 0.6 $\Hz$ without.} In contrast, our network runs at a quicker inference rate of \textit{37 $\Hz$} (27 ms) to perform the Bayesian update, and 13.2 $\Hz$ (76 ms) to query the map.


We train ConvBKI with the Adam optimizer \cite{AdamOptimizer} at a learning rate of 0.007 for one epoch using the weighted negative log likelihood loss. We initialize the kernel length parameter to $l=0.5 \m$ and train ConvBKI with the last $\mathcal{T}=10$ frames, as we found $10$ frames to optimally balance performance and training time.

\subsection{Ablation Studies}
We perform a series of ablation studies over filter size and 
\textcolor{black}{voxel} resolution of ConvBKI Compound on Semantic KITTI \cite{SemanticKITTI} sequence 8. Sequence 8 is part of the validation set and therefore has not been previously seen by the semantic segmentation network during training. We train and test on a voxel grid with bounds of [-20, -20, -2.6] to [20, 20, 0.6] $\m$ along the (X, Y, Z) axes, where points outside of the voxel grid are discarded and not measured in the results. Average latency of the ConvBKI layer is measured on an NVIDIA RTX 3090 GPU over $100$ repetitions, with standard deviation $< 0.4 \mathrm{ms}$.

First, we study the effect of \textcolor{black}{voxel} resolution on the inference time and mIoU of ConvBKI. We compare ConvBKI with resolutions 0.1, 0.2 and 0.4~$\m$, and a constant filter size $f=5$ for all models. Table~\ref{tab:resolution} indicates that a finer resolution can increase performance, however the segmentation difference between 0.2 and 0.1~$\m$ resolution is marginal at the cost of greater memory and slower inference. For real-time driving applications, this suggests that 0.2~$\m$ resolution may be a strong middle ground. Note that the optimal resolution will vary between applications.
 
 Next, we study the effect of the filter size on inference time and performance. While a larger filter size increases the receptive field of the kernel and potentially improves the predictive capability as a result, filter size also cubically increases computation cost. Therefore, identifying a balance between filter size and computational efficiency is important for real-time application. We study filters of size $f=$ 3, 5, 7, and 9 at a resolution of 0.2~$\m$. Table \ref{tab:Filter_Size} demonstrates that filter sizes can improve segmentation accuracy, however quickly increase run-time. In practice a filter size of 5 or 7 may be optimal, as a filter size of 9 offers little improvement with a large increase in computational cost. 
 
 \begin{table}[t]
\centering
\caption{Ablation study of voxel resolution on Semantic KITTI sequence 8 for compound ConvBKI with filter size $f=5$.}
\scriptsize
\begin{tabular}{|c|c|c|c|c|}
 \toprule
 \textbf{Resolution} & \textbf{mIoU (\%)} & \textbf{Latency ($\mathrm{ms}$)} & \textbf{Mem. (GB)} \\
 \bottomrule
 \hline
 N/A (Input) & 54.6 & n/a & n/a\\
 \hline
 0.4~$\m$ & 58.2 & \textbf{8.5} & \textbf{2.4}\\
 \hline
 0.2~$\m$ & \textbf{59.3} & 11.1 & 2.7\\
 \hline
 0.1~$\m$ & 59.0 & 30.1 & 5.0\\
 \bottomrule
 \end{tabular}
\label{tab:resolution}
\vspace{-2mm}
\end{table}

\begin{table}[t]
\centering
\caption{Ablation study of filter size on Semantic KITTI sequence 8 for compound ConvBKI with resolution $0.2 \m$.}
\scriptsize
\begin{tabular}{|c|c|c|c|}
 \toprule
 \textbf{Filter Size} & \textbf{mIoU (\%)} & \textbf{Latency ($\mathrm{ms}$)}\\
 \bottomrule
 \hline
 N/A (Input) & 54.6 & n/a\\
 \hline
 $f=3$ & 59.0 & \textbf{9.5}\\
 \hline
 $f=5$ & 59.3 & 11.1\\
 \hline
 $f=7$ & 59.5 & 13.5\\
 \hline
 $f=9$ & \textbf{59.6} & 17.6\\
 \bottomrule
 \end{tabular}
\label{tab:Filter_Size}
\vspace{-4mm}
\end{table}

 \begin{table}[t]
\centering
\caption{Semantic results on KITTI Odometry sequence 15 \cite{KITTI_Seq_15}.}
\resizebox{0.48\textwidth}{!}{
\begin{tabular}{l|cccccccc|c}
\multicolumn{1}{l}{\bf Method}&

\cellcolor{sbuildingColor}\rotatebox{90}{\color{white}Building} &
\cellcolor{sroadColor}\rotatebox{90}{\color{white}Road} &
\cellcolor{svegetationColor}\rotatebox{90}{\color{white}Vege.} &
\cellcolor{ssidewalkColor}\rotatebox{90}{\color{white}Sidewalk} & 
\cellcolor{sothervehicleColor}\rotatebox{90}{\color{white}Car} & 
\cellcolor{strafficsignColor}\rotatebox{90}{\color{white}Sign} &
\cellcolor{sfenceColor}\rotatebox{90}{\color{white}Fence} &
\cellcolor{spoleColor}\rotatebox{90}{\color{white}Pole} &
\rotatebox{90}{\bf Average} \\

\hline 

\vspace{-2mm} \\
Segmentation \cite{Deep_Dilated_CNN} & 92.1 & 93.9 & 90.7 & 81.9 & 94.6 & 19.8 & 78.9 & 49.3 & 75.1\\
\bottomrule
\vspace{-2mm} \\
Yang et al. \cite{YangMethod9} & \textbf{95.6} & 90.4 & \textbf{92.8} & 70.0 & 94.4 & 0.1 & \textbf{84.5} & 49.5 & 72.2\\
BGKOctoMap-CRF \cite{BKIOccupancy} & 94.7 & 93.8 & 90.2 & 81.1 & 92.9 & 0.0 & 78.0 & 49.7 & 72.5\\
S-CSM \cite{MappingSBKI} & 94.4 & 95.4 & 90.7 & 84.5 & 95.0 & 22.2 & 79.3 & 51.6 & 76.6\\
S-BKI \textcolor{black}{(fine)} & 94.6 & 95.4 & 90.4 & 84.2 & 95.1 & \textbf{27.1} & 79.3 & 51.3 & 77.2\\

\bottomrule
\vspace{-2mm} \\
\textcolor{black}{S-BKI (0.2m)} & 92.6 & 94.7 & 90.9 & 84.5 & 95.1 & 21.9 & 80.0 & 52.0 & 76.5\\
ConvBKI Single & 92.7 & 94.8 & 90.9 & 84.7 & 95.1 & 22.1 & 80.2 & 52.1 & 76.6\\
ConvBKI Per Class & 94.0 & 95.5 & 91.0 & 87.0 & 95.1 & 22.8 & 81.8 & 52.9 & 77.5\\
ConvBKI Compound & 94.0 & \textbf{95.6} & 91.0 & \textbf{87.2} & \textbf{95.1} & 22.8 & 81.9 & \textbf{54.3} & \textbf{77.7}\\
\end{tabular}
}

\label{tab:kitti_iou}
\vspace{-4mm}
\end{table}

\begin{table*}[t]
\centering
\caption{Semantic results on Semantic KITTI \cite{SemanticKITTI} validation and test set.}
\resizebox{\textwidth}{!}{
\begin{tabular}{l|l|llccccccccccccccccc|c|c}
{\bf Data Split} & 
\multicolumn{1}{l}{\bf Method}&
\cellcolor{scarColor}\rotatebox{90}{\color{white}Car} &
\cellcolor{sbicycleColor}\rotatebox{90}{\color{white}Bicycle} &
\cellcolor{smotorcycleColor}\rotatebox{90}{\color{white}Motorcycle} &
\cellcolor{struckColor}\rotatebox{90}{\color{white}Truck} & 
\cellcolor{sothervehicleColor}\rotatebox{90}{\color{white}Other Veh.} & 
\cellcolor{spersonColor}\rotatebox{90}{\color{white}Person} &
\cellcolor{sbicyclistColor}\rotatebox{90}{\color{white}Bicyclist} &
\cellcolor{smotorcyclistColor}\rotatebox{90}{\color{white}Motorcyclist} &
\cellcolor{sroadColor}\rotatebox{90}{\color{white}Road} &
\cellcolor{sparkingColor}\rotatebox{90}{\color{white}Parking} &
\cellcolor{ssidewalkColor}\rotatebox{90}{\color{white}Sidewalk} &
\cellcolor{sothergroundColor}\rotatebox{90}{\color{white}Other Gr.} &
\cellcolor{sbuildingColor}\rotatebox{90}{\color{white}Building} &
\cellcolor{sfenceColor}\rotatebox{90}{\color{white}Fence} &
\cellcolor{svegetationColor}\rotatebox{90}{\color{white}Vegetation} &
\cellcolor{strunkColor}\rotatebox{90}{\color{white}Trunk} &
\cellcolor{sterrainColor}\rotatebox{90}{\color{white}Terrain} &
\cellcolor{spoleColor}\rotatebox{90}{\color{white}Pole} &
\cellcolor{strafficsignColor}\rotatebox{90}{\color{white}Sign} &
\rotatebox{90}{\bf Average} \\ \hline 

\vspace{-2mm} \\
\multirow{2}{*}{Val}
& Segmentation \cite{DarkNet} & 91.0 & 25.0 & 47.1 & 40.7 & 25.5 & 45.2 & 62.9 & 0.0 & 93.8 & 46.5 & 81.9 & \textbf{0.2} & 85.8 & 54.2 & 84.2 & 52.9 & 72.7 & 53.2 & 40.0 & 52.8\\
& \textcolor{black}{S-BKI (0.2m)} & 92.6 & 30.3 & 55.3 & 43.1 & 25.0 & 51.9 & 69.9 & 0.0 & 93.6 & 46.8 & 81.9 & 0.1 & 87.9 & 57.5 & 86.0 & 59.8 & 74.0 & 60.0 & 43.2 & 55.7\\
& ConvBKI Single & 92.0 & 29.8 & 57.4 & 44.4 & 25.2 & 53.1 & 72.1 & 0.0 & 93.1 & 45.8 & 80.9 & 0.1 & 88.2 & 57.8 & 86.1 & 61.2 & 74.0 & 59.7 & 44.4 & 56.1\\
& ConvBKI Per Class & 92.6 & 34.5 & 59.2 & 34.6 & 39.4 & 58.6 & 73.5 & 0.0 & 93.0 & 47.2 & 80.9 & 0.1 & 88.4 & 58.3 & 86.4 & 61.7 & 74.2 & 58.4 & \textbf{47.4} & 57.3\\ 
& ConvBKI Compound & \textbf{94.0} & \textbf{37.5} & \textbf{60.0} & 33.3 & \textbf{40.5} & \textbf{59.4} & \textbf{74.4} & 0.0 & 93.3 & 49.0 & 81.2 & 0.1 & 88.5 & 59.5 & 86.8 & 62.2 & 75.0 & 59.9 & 46.5 & \textbf{58.0}\\
& S-BKI (fine) & 93.5 & 33.5 & 57.3 & \textbf{44.5} & 27.2 & 52.9 & 72.1 & 0.0 & \textbf{94.4} & \textbf{49.6} & \textbf{84.0} & 0.0 & \textbf{88.7} & \textbf{59.6} & \textbf{86.9} & \textbf{62.5} & \textbf{75.3} & \textbf{63.6} & 45.1 & 57.4\\

\bottomrule
\multirow{2}{*}{Test}
& Segmentation \cite{DarkNet} & 82.4 & 26.0 & 34.6 & 21.6 & 18.3 & 6.7 & 2.7 & 0.5 & 91.8 & 65.0 & 75.1 & 27.7 & 87.4 & 58.6 & 80.5 & 55.1 & 64.8 & 47.9 & 55.9 & 47.5\\
& \textcolor{black}{S-BKI (0.2m)} & \textbf{84.0} & 28.5 & 39.9 & 25.2 & 19.7 & 7.9 & 3.3 & 0.0 & 92.3 & \textbf{67.5} & 76.5 & 28.5 & 89.1 & 61.5 & 82.3 & 61.6 & 66.5 & 55.3 & 64.4 & 50.2\\
& ConvBKI Compound &  83.8 & \textbf{32.2} & \textbf{43.8} & \textbf{29.8} & \textbf{23.2} & 8.3 & 3.1 & 0.0 & 91.4 & 62.6 & 75.2 & 27.5 & 89.1 & 61.6 & 81.6 & 62.5 & 65.2 & 53.9 & 63.0 & 50.4\\ 
& S-BKI \textcolor{black}{(fine)} & 83.8 & 30.6 & 43.0 & 26.0 & 19.6 & \textbf{8.5} & \textbf{3.4} & 0.0 & \textbf{92.6} & 65.3 & \textbf{77.4} & \textbf{30.1} & \textbf{89.7} & \textbf{63.7} & \textbf{83.4} & \textbf{64.3} & \textbf{67.4} & \textbf{58.6} & \textbf{67.1} & \textbf{51.3}\\
\bottomrule
\end{tabular}
}
\label{tab:kitti_results}
\vspace{-4mm}
\end{table*}

\subsection{KITTI Dataset}
Following \cite{MappingSBKI}, we evaluate on the KITTI dataset \cite{KITTI_Odometry} with semantically labeled images from \cite{KITTI_Seq_15} as there exist semantic mapping benchmarks for comparison. We follow the same process as \cite{MappingSBKI}, where depth is estimated from ELAS \cite{ELAS_Depth}, pose is estimated from ORB-SLAM \cite{ORBSLAM}, and semantic labels are estimated from the deep network dilated CNN \cite{Deep_Dilated_CNN}. We compare against a CRF-based semantic mapping system \cite{YangMethod9}, BGKOctoMap-CRF \cite{BKIOccupancy, MappingSBKI}, S-BKI \cite{MappingSBKI}, and S-CSM \cite{MappingSBKI}, which all have previously established baselines. Images are projected to 3D and updated by ConvBKI with bounds [-40, -40, -5.0] to [40, 40, 5.0] $\m$, a resolution of 0.2 $\m$, and a filter size of $f=5$. Semantic segmentation performance is calculated for all image points within 40 $\m$ of the ego vehicle. 

Table \ref{tab:kitti_iou} details the performance of the semantic segmentation input, each baseline, and each variation of ConvBKI. We find that the optimized ConvBKI Single performs slightly \textcolor{black}{better than its direct comparison S-BKI (0.2m). Likewise, more expressive kernels increase performance as ConvBKI Compound has a higher mIoU than ConvBKI Per Class, which has a higher mIoU than ConvBKI Single, as expected. ConvBKI Compound with 0.2 $\m$ resolution and discretization can also improve upon the mIoU and latency of S-BKI (fine), which has a finer 0.1 $\m$ resolution without discretization. The improvement of ConvBKI Compound is due to optimization, a more expressive kernel, and hardware acceleration on GPU.} 

 
\subsection{Semantic KITTI}

We perform quantitative analysis on the Semantic KITTI \cite{SemanticKITTI} data set. We train a ConvBKI filter with bounds [-40, -40, -2.6] to [40, 40, 2.6] $\m$, resolution 0.2 $\m$, and filter size 5 following the results of the ablation studies. We compare again against S-CSM, and S-BKI \cite{MappingSBKI} with Darknet53-kNN \cite{DarkNet} as semantic segmentation input. For evaluation we increase the bounds to [-50, -50, -2.6] to [50, 50, 2.6] $\m$ and assign points outside the map to the semantic segmentation network predictions, since only local points within the boundaries are updated in the map.  

Table \ref{tab:kitti_results} details the results of ConvBKI trained on the validation set of Semantic KITTI, compared to the baselines and input semantic segmentation network on both the validation and test set. \textcolor{black}{Similar to Table \ref{tab:kitti_iou}, on the validation set, ConvBKI Single achieves a higher mIoU than un-optimized S-BKI (0.2m) and has a lower mIoU than more expressive ConvBKI Per Class, which has a lower mIoU than ConvBKI Compound. ConvBKI Compound achieves a higher mIoU than S-BKI (0.2m) on both the validation and test set; however has a lower mIoU than S-BKI (fine) on the test set. The discrepancy is likely due to the combination of a difference in resolution, variation in the test and validation set, and threshold tuning of S-BKI \cite{MappingSBKI}.}


\textcolor{black}{Overall, ConvBKI Compound achieves a higher mIoU than direct comparison S-BKI (0.2m) on all data sets due to optimization and a more expressive kernel. While S-BKI (fine) at a finer 0.1 $\m$ resolution without discretization achieves higher performance on the test set of Semantic KITTI, ConvBKI Compound achieves higher mIoU on the other two data sets with lower latency. For a voxel grid with bounds [-40, -40, -2.6] to [40, 40, 2.6] $\m$, resolution 0.2 $\m$, and filter size 5, ConvBKI updates the map at 37 $\Hz$ and queries local voxels from the global map at 13.2 $\Hz$. In contrast, S-BKI (fine) reports an inference rate of 0.6 $\Hz$.}



\begin{figure}[t]
    \centering
    \begin{subfigure}[t]{0.4\textwidth}
        \centering
        \includegraphics[width=\textwidth]{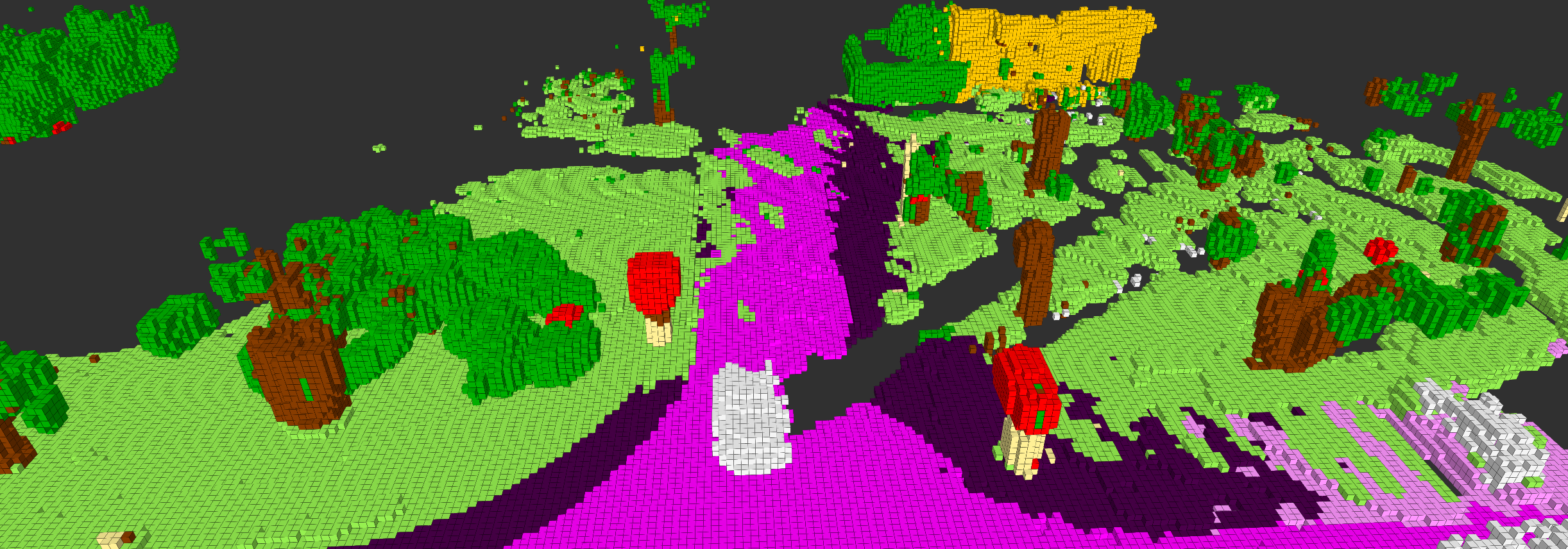}
        \label{fig:bki_raw}
    \end{subfigure}
    \\[-2ex]
    \begin{subfigure}[t]{0.4\textwidth}
        \centering
        \includegraphics[width=\textwidth]{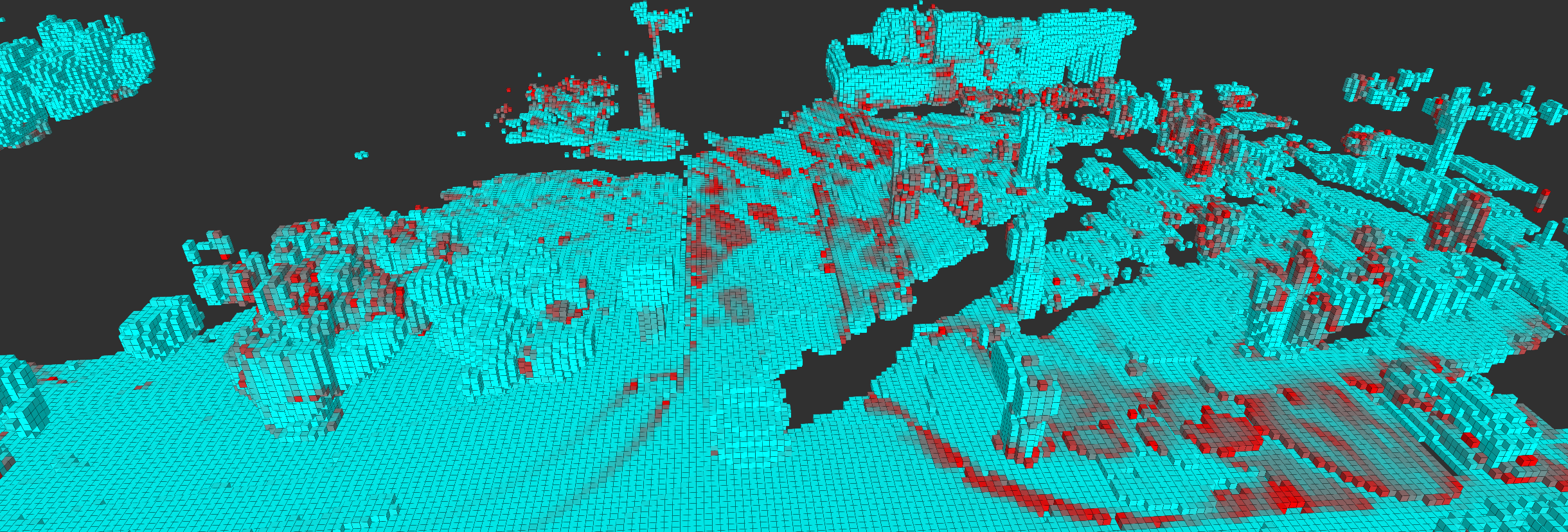}
        \label{fig:bki_variance}
    \end{subfigure}
    \\[-2ex]
    \begin{subfigure}[t]{0.4\textwidth}
        \centering
        \includegraphics[width=\textwidth]{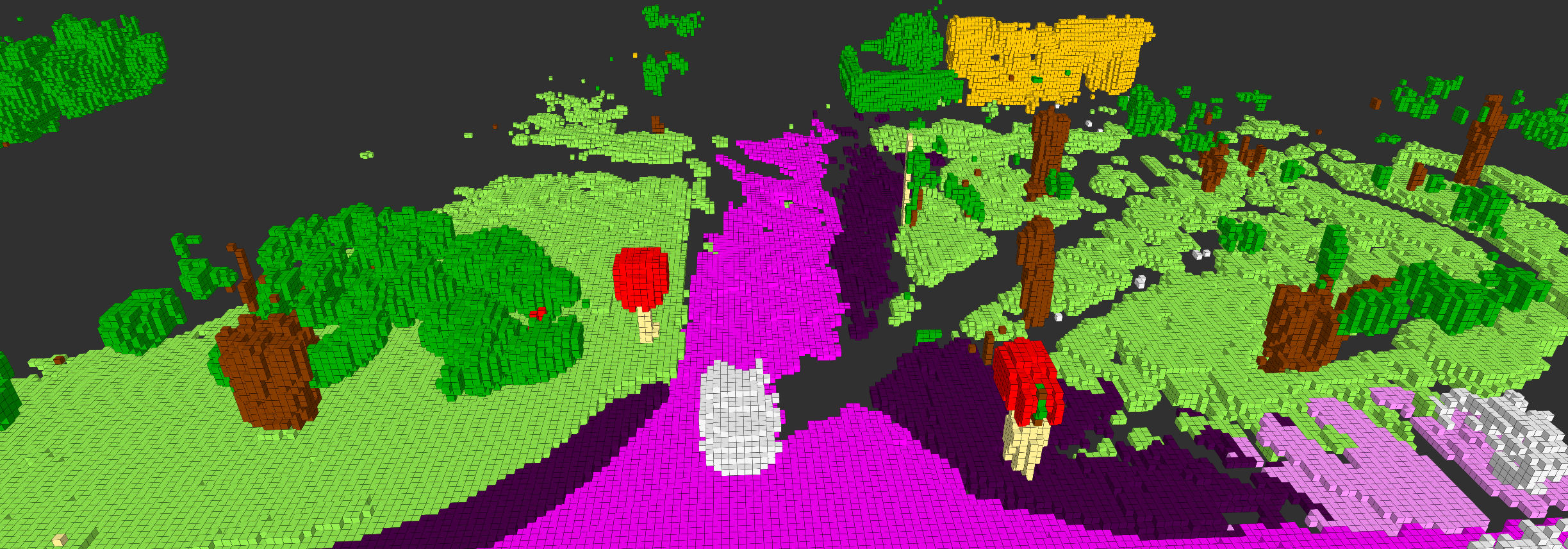}
        \label{fig:bki_filtered}
    \end{subfigure}
    \caption{Example map produced by ConvBKI Compound on the validation set of Semantic KITTI. The expected semantic map is shown in the top image, and the variance is shown in the middle, where red indicates high variance and blue indicates low variance. Removing voxels with high variance or uncertainty (e.g., $\mathcal{V}[\alpha^c_*]$ > 0.01) improves the quality of the robotic map. 
    } 
    \label{fig:SemanticMap}
\label{fig:bki_map}
\vspace{-4mm}
\end{figure}

\subsection{Qualitative Results}
Lastly, we present qualitative results illustrating the distributions learned by the ConvBKI layer, and the generated global map. A video of online mapping \textcolor{black}{can} be found in the supplementary material. 

We include an example map in Fig.~\ref{fig:bki_map} of the Semantic KITTI validation set produced by ConvBKI Compound. The top image demonstrates the expected semantic label produced by the network. As can be seen, there is still noise present, especially around the road. Removing voxels with high variance \textcolor{black}{calculated by Eq. \eqref{eq:Variance}} yields the bottom image, which is improved qualitatively.

\begin{figure}[t]
    \centering
    
    \begin{subfigure}[t]{0.15\textwidth}
        \centering
        \includegraphics[width=\textwidth]{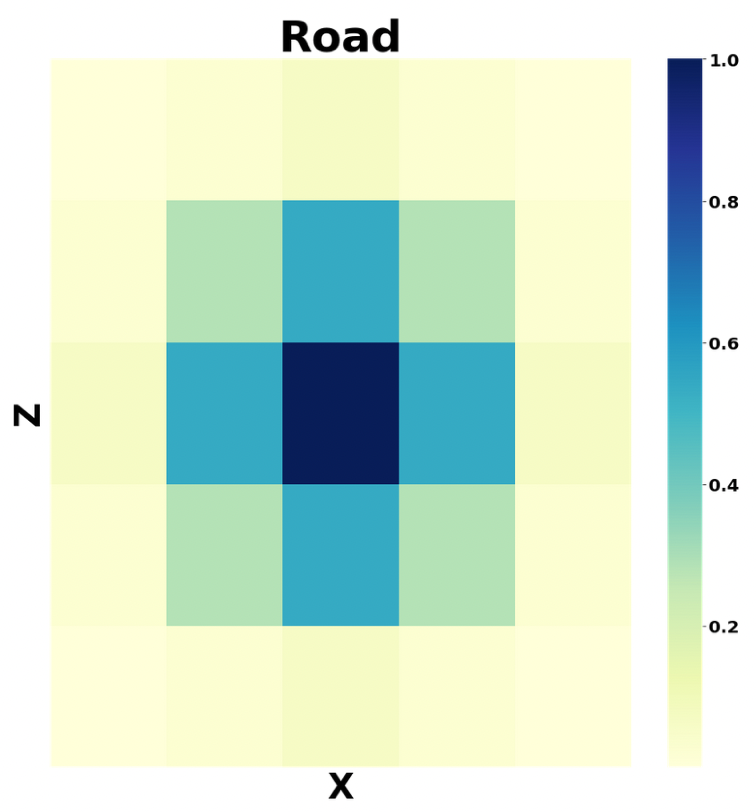}
        \label{fig:sing_road}
    \end{subfigure}
    \hfill
    \begin{subfigure}[t]{0.15\textwidth}
        \centering
        \includegraphics[width=\textwidth]{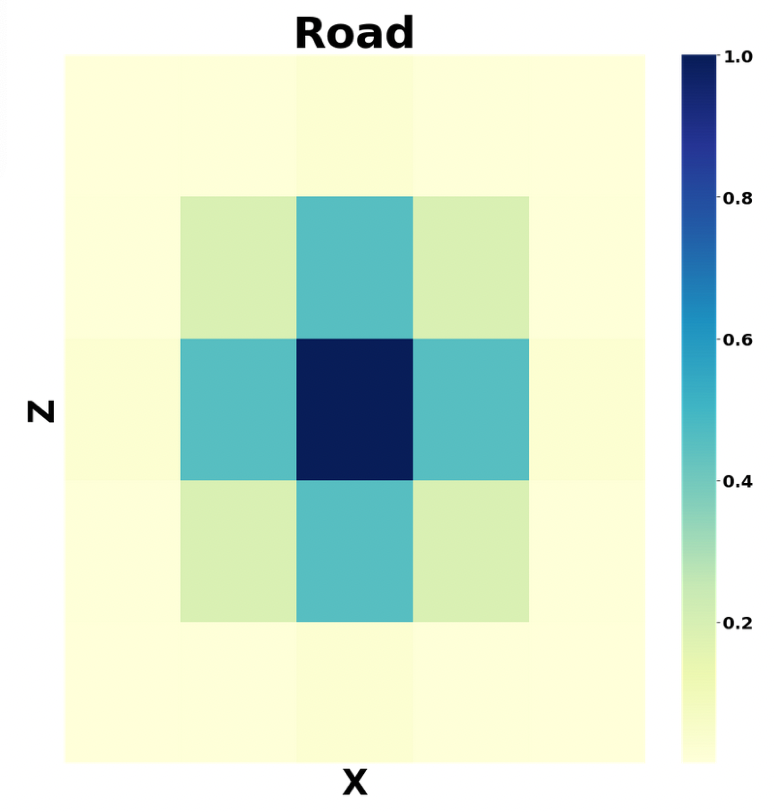}
        \label{fig:pc_road}
    \end{subfigure}
    \hfill
    \begin{subfigure}[t]{0.15\textwidth}
        \centering
        \includegraphics[width=\textwidth]{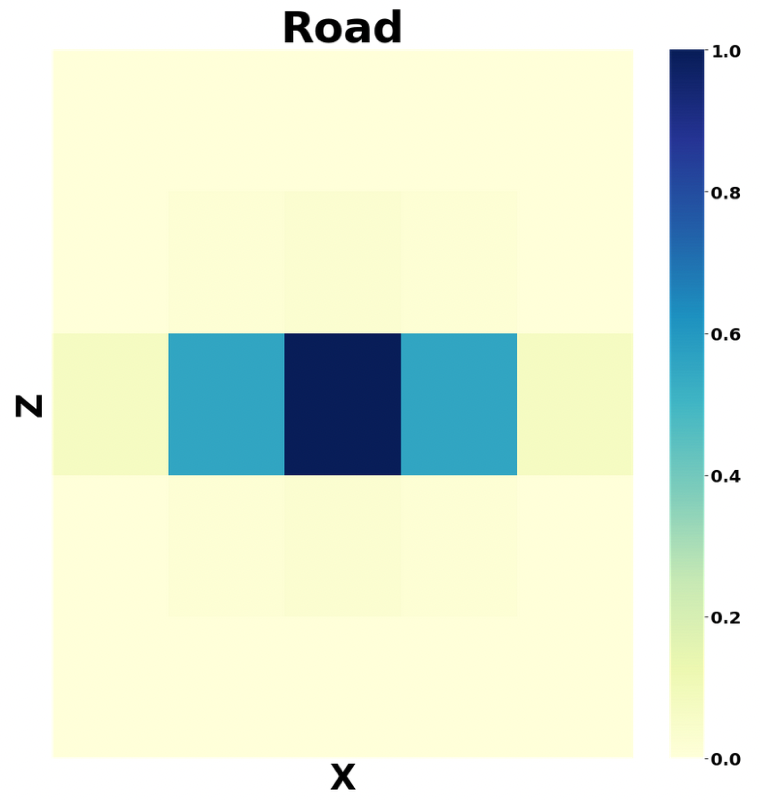}
        \label{fig:comp_road}
    \end{subfigure}
    \hfill

    \begin{subfigure}[b]{0.15\textwidth}
        \centering
        \includegraphics[width=\textwidth]{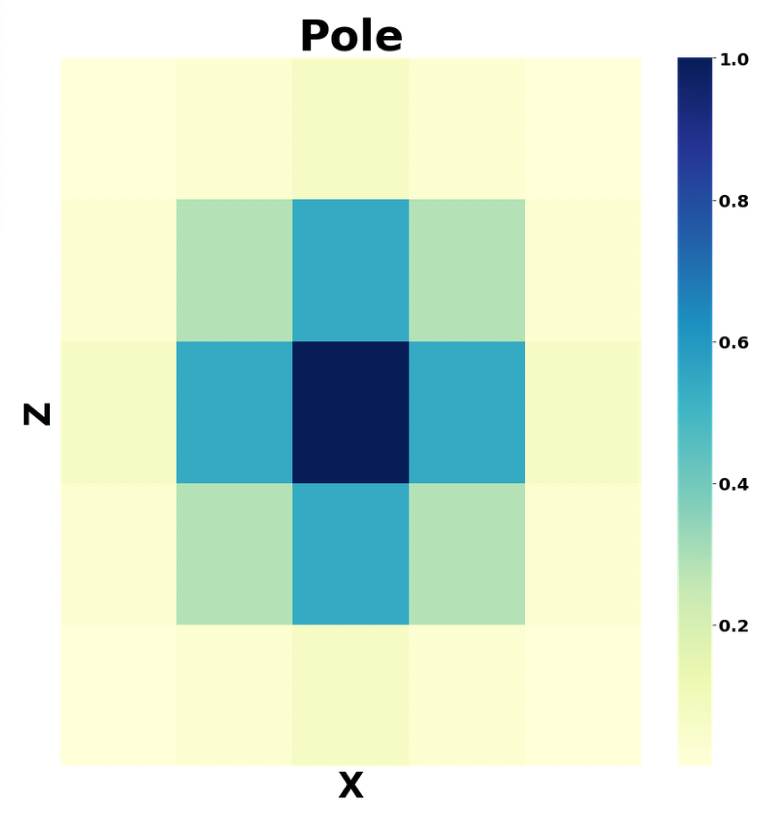}
        \caption{Single}
        \label{fig:sing_pole}
    \end{subfigure}
    \hfill
    \begin{subfigure}[b]{0.15\textwidth}
        \centering
        \includegraphics[width=\textwidth]{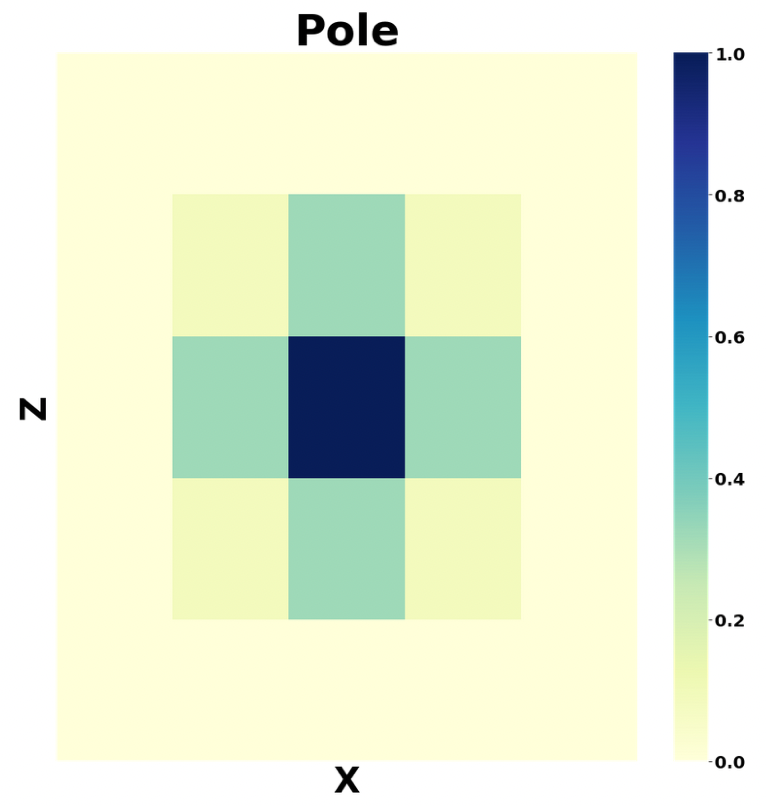}
        \caption{Per Class}
        \label{fig:pc_pole}
    \end{subfigure}
    \hfill
    \begin{subfigure}[b]{0.15\textwidth}
        \centering
        \includegraphics[width=\textwidth]{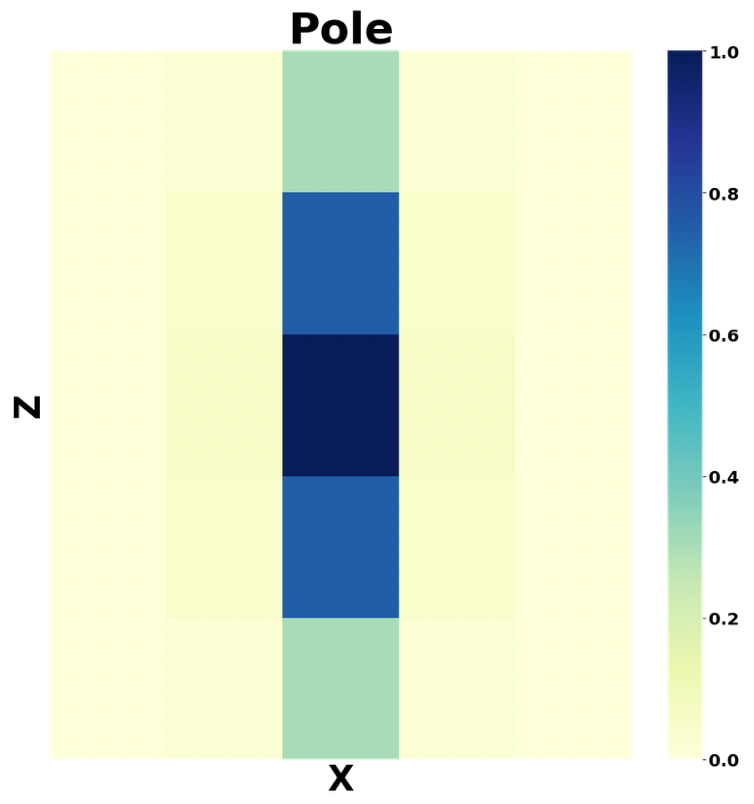}
        \caption{Compound}
        \label{fig:comp_pole}
    \end{subfigure}
    \caption{Illustration of kernels learned by ConvBKI on the road and pole semantic classes, plotted at $\Delta Y=0$. Adding degrees of freedom increases expressivity by allowing the kernel to learn class-specific geometry.} 
    \label{fig:KernelDemo}
\vspace{-4mm}
\end{figure}

Fig.~\ref{fig:KernelDemo} demonstrates the kernels learned by variations of the ConvBKI layer for single, per class, and compound kernels. Each variation of ConvBKI improves potential semantic-geometric expressiveness. ConvBKI Single learns a spherical semantic-geometric distribution shared between all classes. However, semantic classes do not share the same geometry in the real world. ConvBKI Per Class adds the capability to learn a unique distribution for each semantic category, but is still limited by spherical geometry. ConvBKI Compound learns a 3D ellipsoid which can be more expressive for classes such as pole or road.

%% file: Conclusion.tex
\section{Conclusion}
In this paper, we introduced a differentiable 3D semantic mapping algorithm which combines reliability and trustworthiness of classical probabilistic mapping algorithms with the efficiency and differentiability of modern neural networks. We demonstrated that our network can achieve improved results compared to previous 3D mapping approaches, with real-time inference rates. For future work we intend to investigate \textcolor{black}{the ability of ConvBKI to extend} to other data sets and real world mobile robots, \textcolor{black}{propagation of dynamic objects within the BKI framework~\cite{unnikrishnan2022dynamic}, and other methods to accelerate mapping}.